\documentclass[lettersize,journal]{IEEEtran}
\usepackage{amsmath,amsfonts}
\DeclareMathOperator{\sign}{sign}
\usepackage{url}
\usepackage{algorithm,algpseudocode}
\usepackage{array}
\usepackage[caption=false,font=normalsize,labelfont=sf,textfont=sf]{subfig}
\usepackage{textcomp}
\usepackage{stfloats}
\usepackage{url}
\usepackage{verbatim}
\usepackage{graphicx}
\usepackage{multirow}
\usepackage{cite}
\usepackage[capitalise]{cleveref}
\graphicspath{ {./picture/} }
\begin{document}

\title{MSTFormer: Motion Inspired Spatial-temporal Transformer with Dynamic-aware Attention for long-term Vessel Trajectory Prediction}

\author{Huimin Qiang, Zhiyuan Guo, Shiyuan Xie, Xiaodong Peng*
        % <-this % stops a space
% \thanks{The authors are with Key Laboratory of Electronics and Information Technology for Space System, National Space Science Center,
% Chinese Academy of Sciences, Beijing 100190, China;  University of Chinese Academy of Sciences, Beijing 100049, China }% <-this % stops a space
% % qianghuimin21@mails.ucas.ac.cn, guoozy@126.com, xieshiyuan21@mails.ucas.ac.cn, Pxd@nssc.ac.cn
% \thanks{Corresponding author: Xiaodong Peng, Email:Pxd@nssc.ac.cn}
}

% \markboth{IEEE TRANSACTIONS ON INTELLIGENT TRANSPORTATION SYSTEMS}%

\maketitle
\begin{abstract}
% While considering the spatial and temporal characteristics of the vessel, identifying the navigation state of the ship and incorporating the dynamics knowledge into the model is critical for achieving accurate trajectory prediction.
Incorporating the dynamics knowledge into the model is critical for achieving accurate trajectory prediction while considering the spatial and temporal characteristics of the vessel.
% However, existing methods rarely consider the underlying dynamics knowledge, or even ignore the vessel's own motion and use machine learning algorithms to predict the trajectories directly.
However, existing methods rarely consider the underlying dynamics knowledge and directly use machine learning algorithms to predict the trajectories.
Intuitively, the vessel's motions are following the laws of dynamics, e.g., the speed of a vessel decreases when turning a corner. 
Yet, it is challenging to combine dynamic knowledge and neural networks due to their inherent heterogeneity.
% Since dynamic knowledge and neural networks are inherently heterogeneous structures, it is challenging to combine them.
Against this background, we propose MSTFormer, a motion inspired vessel trajectory prediction method based on Transformer.
The contribution of this work is threefold.
First, we design a data augmentation method to describe the spatial features and motion features of the trajectory.
Second, we propose a Multi-headed Dynamic-aware Self-attention mechanism to focus on trajectory points with frequent motion transformations.
Finally, we construct a knowledge-inspired loss function to further boost the performance of the model.
Experimental results on real-world datasets show that our strategy not only effectively improves long-term predictive capability but also outperforms backbones on cornering data.
The ablation analysis further confirms the efficacy of the proposed method. 
To the best of our knowledge, MSTFormer is the first neural network model for trajectory prediction fused with vessel motion dynamics, providing a worthwhile direction for future research.
The source code is available at https://github.com/simple316/MSTFormer.
\end{abstract}


\section{Introduction}
\IEEEPARstart{M}{aritime} transportation is the foundation of international trade and the global economy---more than 80$\%$ of the world's products are transported by sea.
Despite the disruption of COVID-19, the decline in maritime traffic in 2020 is still higher than expected \footnote{https://unctad.org/webflyer/review-maritime-transport-2021}.
With the gradual liberalization of the epidemic and the increasing demand for maritime transportation, the issue of vessel safety and security has received increasing attention from industry and academia \cite{ashraf2022survey}.
Among them, accurate vessel trajectory prediction plays a critical role in collision avoidance \cite{%lei2020mining,
murray2019ais}, abnormal trajectory detection \cite{%perera2012maritime,
liu2022stmgcn}, navigation safety guarantee \cite{liu2022deep}, port management \cite{liang2021mvffnet}, etc.
\par
With the improvement of sensor accuracy, multi-source data is collected for maritime Situation Awareness (SA) and vessel trajectory prediction assistance, including Automatic Identification System (AIS) data\cite{%pallotta2013vessel,
su2022trajectory}, SAR satellite data\cite{brusch2010ship}, Vessel Monitoring Systems (VMS) data\cite{%brusch2010ship,
guo2018trajectory}, long-Range Identification and Tracking (LRIT) data\cite{alessandrini2018estimated}, etc.
According to this review\cite{zhang2022vessel}, 51 of the 57 studies are based on AIS data for vessel trajectory prediction; this indicates a growing academic interest in exploring new approaches to maritime traffic analysis based on AIS data.
% Among them, AIS transponders are widely equipped on vessels to obtain static and dynamic information for use by the Vessel Traffic Service (VTS).
Besides, AIS transponders are widely equipped on vessels, and a large amount of static and dynamic information is acquired to provide data support for the study.
However, AIS data usually include significant data flaws and poor data consistency, which pose many problems for data processing.
At the same time, existing methods use cleaned time-series data directly and do not explore the deep dynamic information contained in the data.
Therefore, how to deeply mine the spatial-temporal information of the trajectory implied in the AIS data is still an open problem.
% Therefore, the accurate prediction of vessel trajectories is still a challenge due to issues of data quality and inadequate data use.
% With the increasing sensor accuracy, there is a growing academic interest in exploring new methods for maritime traffic analysis based on AIS data\cite{su2022trajectory,pallotta2013vessel,xiao2017maritime,xiao2020big}.
\par
Many researchers have devoted their efforts to predicting vessels' trajectory from AIS data \cite{su2022trajectory,xiao2017maritime,xiao2020big}.
The existing vessel trajectory prediction methods can be classified into four categories: simulation-based, statistics-based, deep learning-based, and hybrid-based.
Generally, simulation-based methods simulate vessel behavior by constructing a motion model\cite{last2014comprehensive}, but their predictions are unreliable when the trajectory is sparse\cite{zhang2022vessel}.
Statistics-based methods usually search for matches or build statistical models such as nearest neighbor search \cite{hexeberg2017ais,dalsnes2018neighbor}, Markov chains \cite{guo2018trajectory}, filters \cite{lian2019research}, etc. This method is beneficial for long-term trajectory prediction but is computationally expensive and sensitive to parameters.
Deep learning-based methods abstract the data into high-level features by overlaying multiple non-linear layers to learn the complex dependencies in the trajectory samples \cite{zhang2022vessel}.
While deep learning-based methods have shown fast prediction and strong generalization abilities, they are not empirically suitable for predicting medium- and long-term trajectories and struggle to effectively capture long-term data dependencies \cite{xiao2020big}.
%Although it has fast prediction speed and good generalization ability, it is experimentally proven that it is not suitable for medium- and long-term trajectory prediction and cannot extract long-term data dependence \cite{xiao2020big}.
% In addition, the pure deep learning approach will have reduced accuracy in predicting the trajectory of corners \cite{gao2021novel}.
In addition, pure deep learning-based methods are less accurate in predicting the trajectory of corners \cite{gao2021novel}.
%Some scholars have combined the above models to make the method not only have the efficient computation and generalization ability of neural networks, but also make it achieve good performance in longer trajectory predictions.
\par
% To solve these problems, it is natural to consider combining several models for vessel trajectory prediction.
% The hybrid methods combine deep learning with other methods to improve the performance in predicting longer trajectories. 
To solve these problems of predicting longer trajectories, it is natural to consider hybrid methods that combine deep learning with other methods.
There are two mainstream ways of hybrid methods.
The first is to cluster all historical trajectories, design cluster classifiers, and train their own local behavior network for each cluster \cite{murray2020dual,murray2019ais}.
% , then classify the trajectory and then use the local neural network it belongs to for prediction when a new trajectory comes. to use deep learning to predict the speed and direction and builds a dynamics model to predict the location; 
The second is to extract the typical waterway and use it to correct the prediction results of the neural network \cite{xiao2020big,sang2016cpa,last2019interactive}.
%The main idea of all these combinations is to use the existing or extracted knowledge to assist the network prediction, however, this approach is restrictive
% From a high perspective, these methods used existing or extracted knowledge to help neural networks predict trajectories.
% However, this separation of knowledge from the learning process of the neural network limits the ability of deep learning to extract knowledge of the underlying dynamics.
From a higher dimensional perspective, these methods use pre-existing or extracted knowledge to assist neural networks in predicting trajectories.
However, the separation of knowledge from the network's learning process restricts the potential of deep learning to extract the underlying dynamics knowledge.
In summary, to overcome the above issue and achieve accurate vessel trajectory prediction performance for long-term trajectory, it is necessary to handle three important problems as follows:
\begin{enumerate}
% \item{How to deeply extract effective spatial-temporal information from massive historical AIS data?}
% \item{How to ensure the construction of trajectory long-range dependencies while improving the computational efficiency?}
% \item{How to reliably integrate physical knowledge into the learning process of neural networks and improve its generalization ability?}
\item{How to deeply extract effective spatial-temporal information from massive historical AIS data?}
\item{How to keep a focus on navigation state changes while ensuring the long-range dependence of trajectories?}
\item{How to reliably integrate dynamic knowledge into the learning process of neural networks and improve its comprehension?}
\end{enumerate}
\par
To achieve this goal, we propose MSTFormer, a motion inspired spatial-temporal transformer for long-term vessel trajectory prediction. 
%novel Spatio-Temporal Transformer network for long-term Vessel Trajectory Prediction based trajectory prediction strategy
MSTFormer extracts the spatial Characteristics and simple dynamic knowledge from the Augmented Trajectory Matrix (ATM), guarantees long-term dependence through the Dynamic-aware Self-attention mechanism and is coupled with a knowledge inspired loss function. 
% exploits the long-range time dependency by dynamic-aware attention layers
The main contributions of this paper are summarized as follows:
\begin{enumerate}
\item{We propose a novel data augmentation approach based on the AIS trajectory point information.
% (i.e., longitude, latitude, sog, cog, heading). 
A series of matrices are constructed based on the adjacent trajectory information, incorporating basic dynamics knowledge.}
\item We develop a Multi-head Dynamic-aware Self-attention mechanism that focuses on capturing changes in the vessel's motion and building long-term dependence on the trajectory.
% The dynamic-ware attention layer successfully model the trajectory long-range dependencies and improves the computational efficiency.
\item We design a knowledge inspired loss function based on prediction correction.
% where the predicted trajectory points are derived from $\Delta \text{sog}$ and $\Delta \text{cog}$.
\item{The proposed MSTFormer is assessed on realistic vessel trajectories in the Gulf of Mexico. 
Our method outperforms the previous state-of-the-art approaches in long-term and cornering trajectories.}
\end{enumerate}
\par
The remainder of this paper is organized as follows.
Section~\ref{sec:related work} provides a summary of research on state-of-the-art trajectory prediction methods. 
% Section~\ref{sec:MSTFormer} introduces the data augmentation and problem formulation for predicting vessel trajectory.
The MSTFormer for vessel trajectory prediction is presented in Section~\ref{sec:MSTFormer}.  
The superior performance of our method is demonstrated by comparative experiments and effect analysis in Section~\ref{sec:experimental result}. 
Section~\ref{sec:conclusions} concludes this paper by discussing the research findings and potential directions for future investigation.
\section{Related Works}\label{sec:related work}
% Sensing the current situation and predicting the future state of the vessel is one of the key capabilities to achieving system intelligence.
% In the maritime transportation domain, trajectory prediction approaches have been applied in various applications such as traffic safety and security\cite{alessandrini2018estimated}, collision avoidance\cite{johansen2016ship}, traffic anomaly detection\cite{venskus2021unsupervised}, route planning\cite{uney2019data}, port operation\cite{alizadeh2021prediction}.
% Among them, the trajectory prediction methods can be broadly divided into simulation-based, statistics-based, deep learning-based, and hybrid-based methods. 
% And the latter two methods have developed more rapidly in recent years.
% Additionally, new innovative Transformer-based time series data prediction methods are improving quickly. 
% As a result, we will primarily highlight the three aspects of the trajectory prediction methods below.
In this section, we briefly review deep learning-based and hybrid-based trajectory prediction methods, as they are more relevant to our work.
In addition, transformer-based methods for time series data prediction have attracted substantial attention due to their effectiveness.
We review this line of research in the third subsection.
\subsection{Vessel Trajectory Prediction Using Deep Learning-based Methods}
%Due to the rapid development of computational power and vessel trajectory data collection, deep learning-based trajectory prediction methods have recently made tremendous progress\cite{yan2021emerging}.
%A collection of time-stamped points gathered from AIS devices constitutes the majority of the vessel trajectory data. 
% Thus, tackling the issue of time series prediction may be equated to managing trajectory prediction.
Researchers have used various deep learning models for trajectory prediction, including Recurrent Neural Network (RNN)\cite{capobianco2021deep}, Gated Recurrent Unit (GRU)\cite{ bao2022improved},
% nguyen2018vessel,wang2020vessel,
long Short-Term Memory (LSTM)\cite{liu2022deep}, 
% tang2022model,
and Transformer Neural Network\cite{nguyen2021traisformer}.
Among them, LSTM and its variants have become the mainstream approach for trajectory prediction due to its effectiveness in modeling time series dependency. 
To model the bidirectional dependency, the bidirectional LSTM (Bi-LSTM)\cite{gao2018online} has been introduced, maintaining the relevance between historical and future time-series data.
%The bidirectional structure of Bi-LSTM maintains the relevance between historical and future time-series data better than the single-directional LSTM.
%To tackle this research issue, the authors in \cite{wang2020ship,chen2022fra} discussed an attention-based Bi-LSTM.
Wang and Fu \cite{wang2020ship} proposed an attention-based Bi-LSTM to model multiple dependencies adaptively.
% various focuses on the multiple elements of the Bi-LSTM network's hidden information. 
Moreover, several methods combine bidirectional structure, attention mechanism, and encoder-decoder in various ways\cite{wang2020long,zhang2021bi,sekhon2020spatially}, achieving superior performance.
Besides, other modern network architectures, such as graph convolutional network\cite{liu2022stmgcn} and transformer\cite{nguyen2021traisformer} have also been employed to predict vessel trajectory.
%The outcomes of the experiment show that these strategies work better.
\subsection{Vessel Trajectory Prediction Using Hybrid-base Methods}
%With the growing interest in research based on deep learning methods, studies have found that these methods exhibit poor results on long-term trajectory prediction data\cite{perera2012maritime,xiao2017maritime}.
Deep learning-based methods generally exhibit poor performance in long-term trajectory prediction \cite{xiao2017maritime}.
Therefore, some studies have attempted to use knowledge-aided prediction by combining neural networks with other methods.
Some methods \cite{li2019long,suo2020ship} separate the regions using clustering and then train the regional neural network for prediction.
Instead of clustering region, a hybrid architecture\cite{murray2020dual} with three modules—trajectory clustering, trajectory classification, and trajectory prediction was developed.
Another study\cite{murray2021ais} used hierarchical DBSCAN (HDBSCAN) for the trajectory clustering and bidirectional GRU for classification and prediction.
In addition, there are methods that use extracted waterways or similar historical trajectories to assist neural network predictions.
A novel hybrid framework\cite{xiao2020big} consists of three parts: a neural network predicts sog and cog, corrects cog with the waterway and Particle Filter (PF), and finally calculates position using a dynamical model.
A similarity search prediction model comprising DTW for search and LSTM for spatial distance prediction was explored\cite{alizadeh2021vessel}.
However, these methods use knowledge to assist the network prediction separately.
As a result, the prediction accuracy decreases when the vessel motion changes drastically because these methods do not learn the dynamics themselves.
% As a result, these methods lack the ability to be adaptive to complex data.
\subsection{Recent Advances in Transformer Network}
% \subsection{Recent Advances in Transformer Network for time series prediction}
Transformers' outstanding results\cite{vaswani2017attention} in natural language processing(NLP) has attracted a lot of attention in the time series fields\cite{wen2022transformers}.
For time series modeling, the ability of transformers to capture long-range dependencies and interactions is particularly inspiring, and several Transformer variations have been effectively used for prediction tasks\cite{lim2021temporal,zhou2022fedformer}.
For example, AST trains a sparse Transformer model using a generative adversarial encoder-decoder architecture\cite{wu2020adversarial}.
And to capture temporal correlations of various ranges, Pyraformer adopts a hierarchical pyramidal attention module with a binary tree following route\cite{liu2021pyraformer}.
However, most of these methods are designed for short-term prediction, and the computational efficiency will decrease in the long sequence time-series prediction.
LogTrans integrates sparse bias, a Logsparse mask for long sequences, into the self-attention model to reduce computational complexity\cite{li2019enhancing}.
Similarly, Informer chooses $O(log L)$ dominating queries based on the Query and Key similarities to improve computational efficiency.
Additionally, it develops a generative form decoder to directly provide long-term prediction and prevent accumulated error\cite{zhou2021informer}.
There are also variants of transformers that simultaneously extract spatial and temporal features.
Besides a temporal Transformer block to capture temporal dependencies, Traffic Transformer creates an extra Graph neural network unit to capture spatial dependencies\cite{cai2020traffic}.
Likewise, Spatial-temporal Transformer designs a spatial Transformer block to predict traffic flow\cite{xu2020spatial}.
Spatial-temporal graph Transformer  uses an attention-based graph convolution mechanism\cite{yu2020spatio} for pedestrian trajectory prediction.
However, there is still no variant of Transformer fully exploits the vessel trajectory characteristics.
\section{MSTFormer: Konwledge Inspired Spatial-temporal Transformer}\label{sec:MSTFormer}
\begin{figure*}[!t]
\centering
\includegraphics[width=\textwidth]{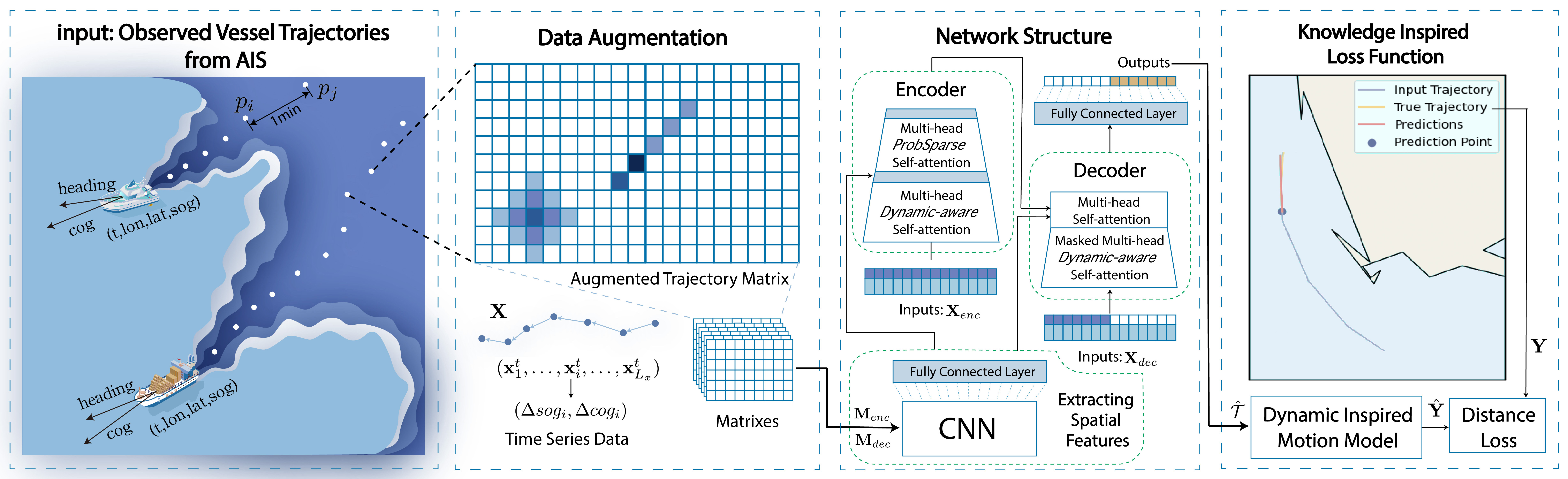}
\caption{Framework of the MSTFormer.}
\label{fig:fig1}
\end{figure*}

In this paper, our goal is to maintain the long-term trajectory dependency and model the spatial-temporal motion features using dynamic knowledge to predict moving vessels' longer positions.
To achieve superior performance, we propose to develop the MSTFormer-based trajectory prediction method, illustrated in Fig. \ref{fig:fig1}. 
The proposed MSTFormer mainly consists of three main components, i.e., Augmented Trajectory Matrix(ATM), network structure with Multi-head Dynamic-aware Self-attention, and knowledge inspired loss function.
Firstly, The Augmented Trajectory Matrix utilizes dynamics knowledge to efficiently represent spatial features efficiently.
Secondly, the network structure with Multi-head Dynamic-aware Self-attention is designed to model long-term trajectory dependency better. 
To extract spatial features from ATMs, a CNN is employed. 
Lastly, the knowledge-inspired loss function calculates the loss using the haversine formula after the motion model utilizes the features extracted by MSTFormer to obtain the predicted position.
% Meanwhile, Multi-head Dynamic-aware Self-attention is proposed to model the long-term trajectory dependency better, and use CNN to extract spatial features from ATMs.
% A motion model uses the features extracted by MSTFormer to obtain the predicted position, then calculate the loss by the haversine formula.

\subsection{Data Augmentation and Problem Formulation}
It is critical to use historical vessel trajectory information to predict future trajectories. We define the trajectory time series and the Augmented Trajectory Matrix to describe the vessel trajectory prediction problem.
\subsubsection{TTS: Trajectory Time Series}
\ 
\newline 
\indent The original data is divided into trajectory segments $Traj_i$ based on time and distance differences, and then uniformly sampled to one point per minute using Cubic spline Interpolation.
The vessel trajectory $Traj_i$ is represented by a series of timestamped trajectory points $p_n=[t_n,lon_n,lat_n,sog_n,cog_n,heading_n]$ with $n\in\{1,2,...,N\}$ obtained by AIS devices, i.e., $Traj_i=\{p_1,p_2,...,p_N\}$ with $N$ is the number of timestamped trajectory points in $T$ and $i\in\{1,2,...,I\}$ with $I$ is the number of trajectory segments. 
Here, $t_n$, $lon_n$, $lat_n$, $sog_n$, $cog_n$, $heading_n$ represent the time stamp, longitude, and latitude, speed over ground, course over ground, heading of the vessel, respectively.
\par
We use the difference of the data as input because the differenced series is smoother, which helps to improve the prediction accuracy.
The sequence data used is represented as $\Delta Traj_i=\{\Delta p_1,\Delta p_2,...,\Delta p_{N-1}\}$, i.e., $\Delta p_m=p_n-p_{n-1}$ with $m\in\{1,2,...,N-1\}$. 
In this work, the input time series of MSTFormer is $\mathbf{X}=\left\{\mathbf{x}_1^t, \ldots, \mathbf{x}_{L_x}^t \mid \mathbf{x}_i^t \in \mathbb{R}^{d_x}\right\}$ and $\mathbf{x}_i^t=[\Delta sog_i,\Delta cog_i]$.
And the input data is split into encoded data and decoded data, which are represented as $\mathbf{X}_{enc}$ and $\mathbf{X}_{dec}$.
The true trajectory is $\mathbf{Y}=\left\{\mathbf{y}_1^t, \ldots, \mathbf{y}_{L_x}^t \mid \mathbf{y}_i^t \in \mathbb{R}^{d_y}\right\}$ and $\mathbf{y}_i^t=[lon_i,lat_i]$.
The historical trajectory point where the prediction begins is called the prediction point and is denoted as $\mathbf{P}=[lon,lat,sog,cog]$.

\subsubsection{ATM: Augmented Trajectory Matrix}
\ 
\newline 
\indent The ATM is constructed from two adjacent trajectory points $p_n$ and $p_{n+1}$. 
In Fig. \ref{fig:fig1}, the position in the center of the matrix is taken as the current position $[lon_n, lat_n]$, and the weight of the center position $ATM[l_{c}]$ is set to $L$. 
According to $cog_n$ and $heading_n$ obtain the representation location $l_{c}$, $l_{h}$ in ATM by the direction ,and assigned as $ATM[l_{h}]=H$ and $ATM[l_{s}]=S \cdot sog_n$ in Algorithm \ref{alg:alg1}, . 
The representation location $l_{n}$ of $p_{n+1}$ in ATM is then calculated based on the location relationship between $p_n$ and $p_{n+1}$. 
And $ATM[l_{n}]$ is Gaussianised in the matrix, where the Gaussian kernel is $G$.
Among them, the constants $L$, $H$, $S$, $sogLine$, and $Grid$  are set to 0.8, 0.2, 0.02, 3, and 0.25 respectively.
The $Grid$ indicates that a grid represents 0.25 KM. 
% The trajectory time series data is different from other time series data in that each trajectory point data contains spatial information and vessel motion state. 
% Therefore, the enhanced method proposed in this paper can better mine spatial information and express simple motion relationships in space.
The ATM proposed in this paper is deemed more appropriate for the analysis of trajectory time series data as it is capable of efficiently extracting spatial features and basic motion states.
% Trajectory time series data is different from other time series data in that each trajectory point data contains spatial information. Therefore, this paper proposed the augmented method can better mine spatial information.

\begin{algorithm}
    \caption{Augmented Trajectory Matrix(ATM)}\label{alg:alg1}
    \begin{algorithmic}[1]
    % \Require{$p_n, p_{n+1}, Grid_{len}, L, H, S, G$ $p_n=[t_n,lon_n,lat_n,sog_n,cog_n,heading_n]$}
    % \Ensure $ATM$
    \Function {getATM}{$p_n, p_{n+1}, L, H, S, G, sogLine, Grid$}
        \State $ATM[l_{c}] \gets L$
        \State $l_{h} \gets [c_x-cos(cog_n),c_y+sin(cog_n)]$
        \State $ATM[l_{h}] \gets H$
        \For{$a=1$ to $sogLine$}
        \State $l_{s} \gets [c_x+acos(cog_n), c_y-asin(cog)_n]$
        \State $ATM[l_{s}] \gets S \cdot sog_n$
        \EndFor
        \State $dis_x \gets Haversine(lon_n,lat_{n+1},lon_n,lat_n)$
        \State $dis_y \gets Haversine(lon_{n+1},lat_n,lon_n,lat_n)$
        \State $\Delta lat \gets lat_{n+1}-lat_{n}$
        \State $\Delta lon \gets lon_{n+1}-lon_{n}$
        \State $l_{n} \gets [\sign  \Delta lat \cdot dis_x / Grid, \sign  \Delta lon \cdot dis_y / Grid]$
        \State $ATM[l_{n}] \gets G$
        \State \Return{$ATM$}
    \EndFunction  
    \end{algorithmic}
\end{algorithm}

\subsubsection{Problem Formulation}
\ 
\newline 
\indent The vessel trajectory prediction problem is a classic time-series prediction task that can be regarded as learning the function $\mathcal{F}$ to estimate the most likely traffic characteristics of a future time period provided historical vessel trajectories.
In this paper, we define the vessel trajectory information on the maritime as the attribute features of the TTS and ATM in the network. 
Thus, the prediction framework in this paper can be given by
\begin{equation}\label{eq:formulation}
\hat{\mathbf{Y}}=\mathcal{F}\left(\mathbf{X}, \mathbf{M}, \mathbf{P} ; \mathbb{C}\right),
\end{equation}
where $\mathcal{F}$ is the MSTFormer network learned from historical data.
$\mathbf{X}$, $\hat{\mathbf{Y}}$ defines the historical trajectory and prediction trajectory, i.e.,$\mathbf{X} = \left\{\mathbf{x}_1^t, \ldots, \mathbf{x}_{L_x}^t \mid \mathbf{x}_i^t \in \mathbb{R}^{d_x}\right\}$, $\hat{\mathbf{Y}} = \left\{\hat{\mathbf{y}}_1^t, \ldots, \hat{\mathbf{y}}_{L_y}^t \mid \hat{\mathbf{y}}_i^t \in \mathbb{R}^{d_y}\right\}$. 
$\mathbf{M}$ denotes the Augmented Trajectory Matrixes by Enhancing raw timing data.
$\mathbf{P}$ is the prediction point.
$\mathbb{C}$ represents a group of network parameters in \cref{eq:formulation}.
\subsection{MSTFormer structure with Dynamic-aware Attention}
This subsection mainly focuses on how to build the structure of MSTFormer to incorporate the knowledge of dynamics to simulate longer spatial-temporal correlations and how to predict vessel trajectory based on MSTFormer.
% We suggest expanding the standard Transformer to a structure that simultaneously focuses on long-trajectory and spatial-temporal properties.
% We first introduce the variant of transformer——informer, designed for efficient long-term time series data prediction problems.
% The core structure of the informer consists of time coding, a Multi-head probsparse attention mechanism, a self-attention distilling operation, and a generative style decoder.
We first introduce the data embedding of MSTFormer, which include position embedding and time embedding.
Then, we introduce the Multi-head Dynamic-aware Self-attention proposed to focus on long-term trajectory dependence and motion state change more effectively.
Finally, The complete network structure of MSTFormer, including encoders with the distilling operation and generative style decoders is performed on TTS and ATM for feature learning.
\subsubsection{Data Embedding}
\ 
\newline 
\indent Transformer has neither recurrence nor convolution, unlike LSTM or RNN.  
Instead, it models the sequence information using position embedding.
For each position $pos$ in the vanilla Transformer, there is the  fixed position embedding
\begin{equation}\label{eq:PE}
P E_{(pos, n)}= \begin{cases}\sin \left(pos / 10000^{2n/d_{model}}\right), & n \% 2=1 \\ \cos \left(pos / 10000^{2n/d_{model}}\right), & n \% 2=0\end{cases}
\end{equation}
where $n$ is the dimension, $d_{model}$ is the feature dimension after token embedding.
This function is possible to encode both absolute and relative positions.
\par
\begin{figure}[htp]
\centering
\includegraphics[width=3in]{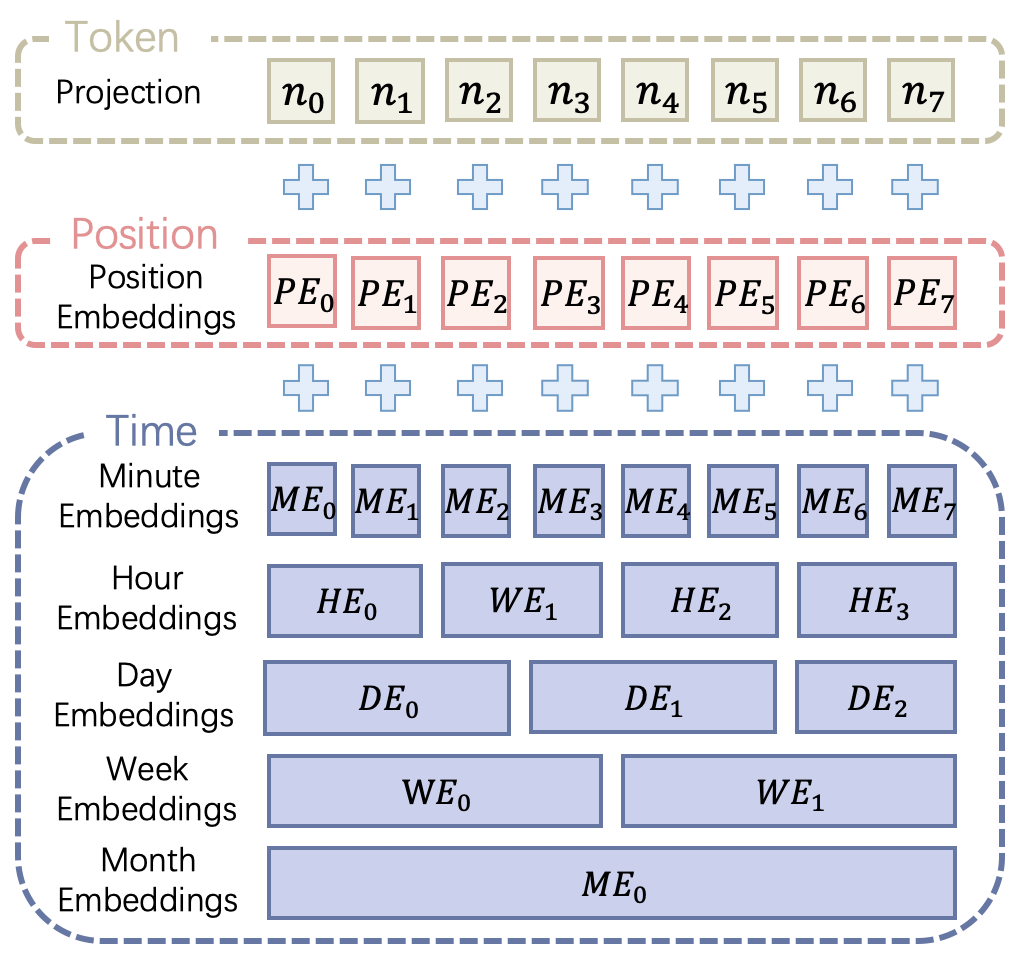}
\caption{data embedding of the MSTFormer.}
\label{fig:fig2}
\end{figure}

In order to capture long-range dependence and manipulate global context, the hierarchical time embedding\cite{zhou2021informer,zhou2022fedformer} is utilized.
This data embedding method has become popular to lessen the impact of the encoder and decoder Query-Key mismatches.
A trainable stamp embeddings $TE_{(pos, i)}$ employs all global time stamp layers and follows the same embedding strategy as $PE_{(pos, i)}$ in \cref{eq:PE}.
Based on the temporal characteristics of the vessel trajectory data, this paper utilizes five layers (minute, hour, day, week, month) of time embedding in Fig. \ref{fig:fig2}. 
This approach can better capture the periodicity of trajectories at different time lengths.
\par
In Fig. \ref{fig:fig2}, we first adjust the dimensionality using a 1-D convolutional filter (kernel width = 3, stride = 1) from the input data $\mathbf{x}_{i}^{t}$ to $d_{model}$-dim vector $\mathbf{u}_i^t$.
Thus, the feeding vector is provided by 
\begin{equation}\label{eq:timecode}
\mathbf{X}_{\text {feed}[i]}^t=\alpha \mathbf{u}_i^t+\operatorname{PE}_{\left(ind,\right)}+\sum_l\left[\operatorname{TE}_{\left(ind,\right)}\right]_l,
\end{equation}
where $ind=L_x \times(t-1)+i$, $i$ is the index of input data and $l$ is the layer number of time embedding, i.e., $i \in\left\{1, \ldots, L_x\right\}$ and $l \in\left\{1, \ldots, L_l\right\}$.
$\alpha$ is a factor to balance the token projection and the embedding.

\subsubsection{Multi-head Dynamic-aware Attention}
\ 
\newline 
\indent 
% To extract the long-term trajectory dependency and reduce the computational complexity simultaneously, it is necessary to robustly and accurately improve the vanilla Multi-head attention mechanism.
Transformer adopts the formulation in the study \cite{vaswani2017attention}, replacing the single attention function \cref{eq:atten} with multi-head attention with $H$ discrete learned projections as \cref{eq:Multiatten}
\begin{equation}\label{eq:atten}
\operatorname{Attention}(Q, K, V)=\operatorname{softmax}\left(\frac{Q K^T}{\sqrt{d_k}}\right) V.
\end{equation}

\begin{equation}\label{eq:Multiatten}
\begin{aligned}
\operatorname{MultiHead}(Q, K, V) =\operatorname{Concat}\left( \ldots, \operatorname{head}_{i},\ldots \right) W^O, \\
\text { where } head_i=\operatorname{Attention}\left(Q W_i^Q, K W_i^K, V W_i^V\right).
\end{aligned}
\end{equation}
\par
\begin{figure}[htp]
\centering
\includegraphics[width=3.3in]{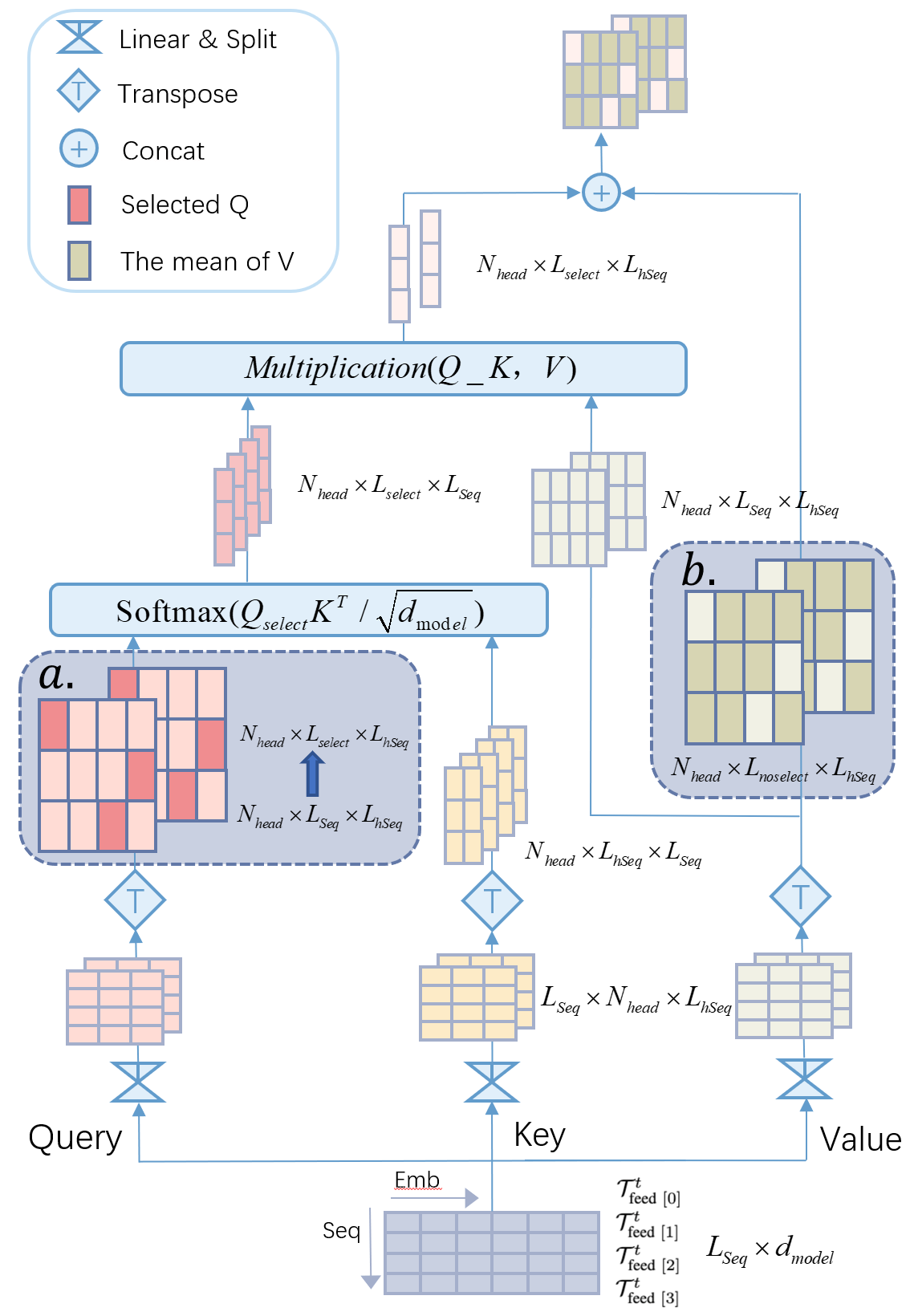}
\caption{Multi-head Dynamic-aware Attention.}
\label{fig:fig3}
\end{figure}

The multiple independent attention allows for parallel computation, which improves computational efficiency.
However, due to the quadratic complexity of sequence length, the long-term prediction problem still requires a large amount of computing.
Our method chooses representative Query in Fig.\ref{fig:fig3}.a and Value in Fig.\ref{fig:fig3}.b to take part in the computation, ensuring accuracy and reducing computational complexity.
\par
In Fig.\ref{fig:fig3}, the input $\mathbf{X}_{\text {feed}} \in \mathbb{R}^{L_{seq} \times d_{model}}$ of the attention block is first linearly projected using $\boldsymbol{w}_Q, \boldsymbol{w}_K,\boldsymbol{w}_V \in \mathbb{R}^{d_{model} \times d_{model}}$, so $Q=\mathbf{X}_{\text {feed}} \cdot \boldsymbol{w}_Q$
Similarly, We can obtain Key and Value.
Then we split $Q$ into multiple heads and transpose it, make $Q \in \mathbb{R}^{N_{head} \times L_{seq} \times L_{hSeq}}$ and $d_{model}=N_{head} \times L_{hSeq}$. 
In each head $i$, we select $M$ representative $\tilde{\boldsymbol{Q}}_{h}^{i}$ from the $\boldsymbol{Q}_{h}^{i}$ by
% We utilize a select operator only to hold the selected $M$ modes
\begin{equation}\label{eq:select_Q}
\tilde{\boldsymbol{Q}}_{h}^{i}=\operatorname{Select}(\boldsymbol{Q}_{h}^{i})=\operatorname{Select}(\boldsymbol{Q}_{h}^{i},\operatorname{Sort}(\mathcal{I}_{h}^{i}(\mathbf{X})),M),
\end{equation}
where$\boldsymbol{Q}=\{\boldsymbol{Q}_{h}^{1}, \ldots, \boldsymbol{Q}_{h}^{N_{head}} \mid \boldsymbol{Q}_{h}^{i} \in \mathbb{R}^{L_{seq} \times L_{hSeq}}\}$, 
$\mathcal{I}=\{\mathcal{I}_{h}^{1}, \ldots, \mathcal{I}_{h}^{N_{head}} \mid \mathcal{I}_{h}^{i} \in \mathbb{R}^{L_{seq}}\}$ and $M=c \cdot \ln L_Q$.
$\mathcal{I}_{h}^{i}$ represents the importance list of $\boldsymbol{Q}_{h}^{i}$ at different positions in the $i$-th head.
It can be obtained by
\begin{equation}\label{eq:dynamic attention all}
\mathcal{I}_{h}^{i}(\mathbf{X})=\mathcal{I}_{pos}(\mathbf{X}_{ind})+ w_{s}\mathcal{I}_{va}(\mathbf{X}_{\Delta sog})+w_{c}\mathcal{I}_{va}(\mathbf{X}_{\Delta cog}),
% \mathcal{I}_{value}(\mathcal{T}_{value})=\ln ( \operatorname{min} (\operatorname{max} (\lvert \mathcal{T}_{value} \rvert - \overline{\mathcal{T}_{value}},0),1))
\end{equation}
\begin{equation}\label{dynamic attention pos}
\mathcal{I}_{pos}(\mathbf{X}_{ind})=\frac{1}{\ln (\mathbf{X}_{ind}+D)}, 
\end{equation}
\begin{equation}\label{eq:dynamic attention value}
\mathcal{I}_{va}(\mathbf{X}_{val})=\ln ( \operatorname{min} (\operatorname{max} (\lvert \mathbf{X}_{val} \rvert - \frac{\sum \lvert \mathbf{X}_{val} \rvert}{\mathbf{X}_{len}},0),1)),
\end{equation}
where $\mathbf{X}_{index}$ is the index of the historical trajectory time series data, $\mathbf{X}_{\Delta sog}$ and $\mathbf{X}_{\Delta sog}$ is the speed difference and course difference between two trajectory points.
$w_{s}$ and $w_{c}$ are used to regulate the proportion of the three influences in the different heads.
$D$ is a constant to Guarantee $\mathcal{I}_{Pos} \in [0,1]$. We can obtain the most important $Q$ by sorting the $\mathcal{I}(\mathbf{X})$. Then the $Q\_K$ is given by
\begin{equation}\label{eq:QK}
\tilde{Q}\_K=\operatorname{softmax}\left(\frac{\tilde{Q} K^T}{\sqrt{d_k}}\right).
\end{equation}
\par
Other than that, we fill the unselected positions with $\overline{V}$, and the attention formulation is 
\begin{equation}\label{eq:calculateV}
\operatorname{Attention}(Q, K, V)=\operatorname{Concat}(\overline{V_{noselect}},\tilde{Q}\_K \cdot V).
\end{equation}

When the data enters the second layer of the MSTFormer, it is impossible to determine its importance based on dynamic knowledge.
In this paper, we use the ProbSparse Self-attention method in Informer \cite{zhou2021informer} above the first layer of MSTFormer.
First, randomly select $\tilde{K}$ from $K$, and the number of $\tilde{K}$ is $c \cdot \ln L_K$. 
Then calculate the sampled $Q\_\tilde{K}=Q\tilde{K}^T$ and
 compute the importance by
\begin{equation}\label{eq:prob attention}
\begin{aligned}
\mathcal{I}(\mathbf{X})=\operatorname{max}(Q\_\tilde{K})-\operatorname{mean}(Q\_\tilde{K}).
\end{aligned}
\end{equation}
%当原始数据进入到第二层网络时，便无法根据motion 知识确定其重要性，本文采用 IN中的p。。。方法，选择Q。首先随机挑选。 个K，计算Q_Ksample
\subsubsection{the Entire Structure of MSTFormer}
\ 
\newline 
\indent The entire structure of the MSTFormer mainly consists of a CNN and a variant of the transformer in Fig. \ref{fig:fig1}.
% While Transformer mostly learns the time series features of the data, CNN primarily learns the spatial features of the trajectory.
The CNN focuses on extracting spatial features from trajectory data and understanding basic dynamics in ATMs. 
In contrast, the Transformer primarily captures the temporal features of the time series data, paying attention to changes in speed and course.
\par
We evenly select a portion of data for the ATMs, after dividing the encoder data and decoder data.
The same three-layer convolutional network is used for the ATMs of the encoder and decoder
\begin{equation}\label{eq:CNN-detail}
Conv=\operatorname{MaxPool3d} (\operatorname{ReLU} (\operatorname{Conv3d} (ATM) ) ),
\end{equation}
\begin{equation}\label{eq:CNN}
\operatorname{CNN}=\operatorname{Linear}(Conv1,Conv2,Conv3),
\end{equation}
where $kernel_{Conv3d}=(1, 3, 3)$ ,$kernel_{MaxPool3d}=(1, 2, 2)$ and the channels remains unchanged in $Conv1$.
$kernel_{Conv3d}=(3, 3, 3)$ ,$kernel_{MaxPool3d}=(3, 2, 2)$ and the channels changed from $1$ to $16$ in $Conv2$.
$kernel_{Conv3d}=(3, 3, 3)$ ,$kernel_{MaxPool3d}=(1, 2, 2)$ and the channels changed from $16$ to the length of time series data in $Conv3$.
% In the first layer, the network mainly extracts the features of individual matrices and makes them smaller in size, reducing the subsequent computation.
In the first layer, the network mainly extracts the shape features in the ATM and learns features about the motion knowledge between vessel speed, course, and next-moment trajectory point position.
% Then, we convolve the depth in the subsequent network, in order to extract the relationship between the vessel motion behavior at different moments.
Next, in the subsequent network, we apply deep convolution to extract the connections between the ship motion behaviors at different moments.
\par
The MSTFormer mainly consists of encoders and decoders. The prediction results can be obtained by
\begin{equation}
    \hat{\mathcal{T}} =\operatorname{Dec}( \operatorname{Enc}(\mathbf{X}_{enc}, \mathbf{M}_{enc};\mathbb{C}) , \mathbf{X}_{dec}, \mathbf{M}_{dec};\mathbb{C}),
\end{equation}
where $\mathbf{X}_{enc}$, $\mathbf{X}_{dec}$, $\mathbf{M}_{enc}$ and $\mathbf{M}_{dec}$ are the encoded data and decoded data split from set $\mathbf{X}$ and set $\mathbf{M}$.
% In the encoder, the input time series data are combined with the tensors obtained from CNN after embedding and attention calculation. 
In the encoder, the input time series data is first processed by embedding and attention, and the resulting tensor is combined with the tensor obtained by CNN.
% The first layer of the encoder incorporates the tensors of the CNN and uses dynamic-aware attention, unlike the other layers in the MSTFormer.
\begin{gather}
\mathcal{A}_{enc} =\operatorname{Norm}(\operatorname{D-atten}(\operatorname{Embedding}(\mathbf{X}_{enc}))+\mathbf{X}_{enc}),\\
\mathcal{C}_{enc} = \operatorname{Norm}(\mathcal{A}_{enc}+\operatorname{CNN}(\mathbf{M}_{enc})), \label{eq:encoder-add-cnn}\\
\operatorname{Enc}(\mathbf{X}_{enc}, \mathbf{M}_{enc};\mathbb{C})=\operatorname{conv1d}(\operatorname{relu}(\operatorname{conv1d}(\mathcal{C}_{enc}))) \label{eq:encoder-final}.
\end{gather}
In the first layer of the encoder, $\operatorname{D-atten}$ and $\operatorname{Norm}$ represent the Multi-head Dynamic-aware Self-attention and layer normalization. 
In the other layers, $\operatorname{D-atten}$ represent Multi-head ProbSparse self-attention, and $\mathcal{C}_{enc} = \operatorname{Norm}(\mathcal{A}_{enc})$ in \cref{eq:encoder-final}.
\par
In MSTFormer, the decoder is implemented by improving the generative structure of Informer\cite{zhou2021informer}.
\begin{gather}
\mathcal{A}_{dec} = \operatorname{Norm}(\operatorname{D-atten}(\operatorname{Embedding}(\mathbf{X}_{dec}))+\mathbf{X}_{dec}),\\
\mathcal{C}_{dec} = \operatorname{Norm}(\mathcal{A}_{dec}+\operatorname{CNN}(\mathbf{M}_{dec})),\\
\mathcal{D}_{dec} = \operatorname{Norm}(\operatorname{atten}(\mathcal{C}_{dec},\mathbf{X}_{enc},\mathbf{X}_{enc})+\mathbf{C}_{dec}),\\
\operatorname{Dec}(\mathbf{X}_{dec}, \mathbf{M}_{dec};\mathbb{C})=\operatorname{conv1d}(\operatorname{relu}(\operatorname{conv1d}(\mathcal{D}_{dec}))), 
\end{gather}
where $\operatorname{atten}$ is the Vanilla attention mechanism. 
This generative style decoder of the MSTFormer has only one layer. 
\subsection{Knowledge Inspired Loss Function}
%数据复原、位置计算、distance loss计算
There are methods that use neural networks to predict the sog and cog of a vessel, and then obtain its latitude and longitude by building a motion physics model\cite{sang2016cpa,xiao2020big}.
The MSTFomer uses $\Delta sog,\Delta cog$ for prediction, and the data series after performing differencing is more stationary, which is beneficial to improve the prediction accuracy.
Additionally, by directly adding the motion model into the loss function to determine the vessel's position, the model can better understand the assessment mechanism.
\par
First, the knowledge inspired loss function performs data recovery on the predicted difference results of the MSTFormer model by
\begin{gather}
(\hat{sog_i}, \tilde{cog}_i)=
{\begin{cases}
    (sog, cog)+\hat{\mathcal{T}}_{i}, & i=1\\
    (\hat{sog}_{i-1}, \hat{cog}_{i-1}) + \hat{\mathcal{T}}_{i}, & i>1
\end{cases}} \label{eq:data_recovery} \\
\hat{cog}={\begin{cases}
        \tilde{cog}+360, & \tilde{cog}<0 \\
        \tilde{cog}-360, & \tilde{cog}>360
    \end{cases}} \label{eq:cog_recovery}
\end{gather}
where $(lon, lat, sog, cog)$ is the prediction point and the $\hat{\mathcal{T}}_{i}$ is network output sequence.
\par
Here, we can use the prediction points $(sog, cog)$ directly to infer the position of the first predicted result, but since the velocity and direction are instantaneous values, there is a bias in the direct calculation.
Therefore, in this paper, the output of the network is considered as a correction of the instantaneous values, and the average speed and average steering between two points are calculated using \cref{eq:data_recovery}.
% Since both sog and cog in the trajectory points are instantaneous values, there is a bias in using them directly to calculate the position.
Meanwhile, for each recovered data $\tilde{cog}_i$, it was corrected using \cref{eq:cog_recovery}.
Then, we use the $\hat{sog}$ and $\hat{cog}$ sequences corrected above and the $lat$ and $lon$ of the prediction point to calculate the predicted location
\begin{gather}
(\hat{lon}_i,\hat{lat}_i)=
{\begin{cases}
    \mathcal{M}(lon,lat,\hat{sog}_i,\hat{cog}_i), & i=1\\
    \mathcal{M}(\hat{lon}_{i-1},\hat{lat}_{i-1},\hat{sog}_i,\hat{cog}_i), & i>1
\end{cases}} 
\end{gather}
where $\mathcal{M}$ is the motion model to obtain the predicted $\hat{lon}$ and $\hat{lat}$.
% Since both sog and cog in the trajectory points are instantaneous values, there is a bias in using them directly to calculate the position.
% Therefore, this paper uses the $\hat{\mathcal{T}}$ sequence to correct the $sog$ and $cog$ in the predicted points in \cref{eq:data_recovery} and uses the corrected values $\hat{sog}$ and $\hat{cog}$ to calculate the true position.
$\mathcal{M}$ is calculated as follows
\begin{gather}
distance=\hat{sog}\times T_{inter} \times N,  \notag  \\
\delta=distance/R(lat),  \notag   \\
\hat{lon} = lon + \arctan(\frac{\sin(\hat{cog}) \sin(\delta)\cos(lat)}{\cos(\delta) - \sin(lat)\sin(\hat{lat})}),     \\
\hat{lat} = \arcsin(\sin(lat)\cos(\delta) + \cos(lat)\sin(\delta)\cos(\hat{cog})), \notag  
\end{gather}
where $T_{inter}$ is time interval between trajectory points. 
Since the Earth is an irregular sphere, use $R(lat)$ to calculate the radius of the Earth at different locations in \cref{eq:earth-redius}.
\begin{gather}\label{eq:earth-redius}
R(lat)=\sqrt{\frac{\left(A^2 \cos lat\right)^2+\left(B^2 \sin lat\right)^2}{(A \cos lat)^2+(B \sin lat)^2}} ,
\end{gather}
where $A$ is the Equatorial radius and $B$ is the Polar radius in WGS-84.
And $A=6378137.0$, $B=6356752.3142$.
\par
Finally, we can use the ‘haversine’ formula to calculate the great-circle distance between the true trajectory and predicted points.
\begin{equation}
\begin{aligned}
    \Delta lat=&lat-\hat{lat},\\
    \Delta lon=&lon-\hat{lon},\\
    a = \sin^2(\Delta lat / 2) +\cos(&\hat{lat}) \cos(lat)\sin^2(\Delta lon / 2), \\
    dis=2R(&\hat{lat})\arcsin(\sqrt{a}),
\end{aligned}
\end{equation}
% \begin{gather*}
% \Delta lat=lat-lat\\
%     \Delta lon=lon-lon\\
%     a = \sin^2(\Delta lat / 2) +\cos(lat) \cos(lat)\sin^2(\Delta lon / 2) \\
%     loss=2R(lat)\arcsin(\sqrt{a})
% \end{gather*}
where $(lon,lat)$ is the true trajectory point, $(\hat{lon},\hat{lat})$ is the predicted point.
The loss can obtain by
\begin{equation}
loss =\frac{\sum\nolimits_{n=1}^N dis}{N}.
\end{equation}
\section{Experimental Results and Analysis}\label{sec:experimental result}
\subsection{Data Processing}
% Although there are some vessel trajectory datasets, due to many articles do not open their datasets and the free data being very dirty, it isn’t easy to access.%数据缺失，衡量尺度不一 
% Therefore, this paper is based on the dataset we cleaned ourselves. 
We downloaded the AIS data from the NOAA (National Oceanic and Atmospheric Administration) \footnote{https://coast.noaa.gov/htdata/CMSP/AISDataHandler/2021/}.
%But as long as the data set is provided, the experiments can be easily confirmed in different marine areas due to the generality of our experimental setup.
% \begin{table}\centering
% \caption {Vessel Type Statistics} \label{tab:vessel-type} 
% \resizebox{80mm}{!}{
% \begin{tabular}{cccll}
% \cline{1-3}
% Vessel Type    & Number & Proportion(\%) &  &  \\ \cline{1-3}
% Pleasure Craft & 4091   & 35.883         &  &  \\
% Towing         & 2520   & 22.103         &  &  \\
% Fishing        & 966    & 8.473          &  &  \\
% Other Type     & 821    & 7.201          &  &  \\
% Sailing        & 809    & 7.096          &  &  \\
% Cargo          & 747    & 6.552          &  &  \\
% Passenger      & 731    & 6.412          &  &  \\
% Tanker         & 372    & 3.263          &  &  \\
% ...            & ...    & ...            &  &  \\ \cline{1-3}
% \end{tabular}}
% \end{table}
% \begin{table}\centering
% \caption {Vessel Type Statistics} \label{tab:vessel-type} 
% \resizebox{!}{!}{
% \begin{tabular}{ccccc}
% \hline
% Vessel Type    & Number & Proportion(\%)   \\ \hline
% Towing         & 958    & 40.286           \\
% Cargo          & 362    & 15.223           \\
% Tanker         & 301    & 12.658           \\
% Other Type     & 289    & 12.153           \\
% Pleasure Craft & 258    & 10.849           \\
% Passenger      & 83     & 3.490            \\
% Sailing        & 58     & 2.439            \\
% Fishing        & 12     & 0.505            \\
% ...            & ...    & ...              \\ \hline
% \end{tabular}}
% \end{table}
The NOAA AIS data consists of 82 different types of vessels, including 11,401 vessels from January 1 to December 31, 2021.
We used data in the longitude range of -98.5 to -80 and the latitude range of 17 to 31, which is the area of the Gulf of Mexico.
The raw data have various problems, including non-uniformity of the measurement scale, missing data, and imbalanced classes.
Therefore, we performed necessary data preprocessing procedures, including cleaning abnormal data, data interpolation, data sampling, and data standardization.
% The largest number of vessels is pleasure crafts, which occupy 35.883\% of the total number of vessels, with a serious imbalance in the number of different types of vessels.
% The experimental results will be influenced by the type of vessel if the raw data is used directly because different vessels travel in different ways.
% At the same time, the raw data have various problems such as non-uniformity of the measurement scale and missing data. 
% For example, some $cog \in [0,360]$, some $cog \in [-180,180].$ %the direction range of some vessels is 0 to 360, and some are -180 to 180. 
% There are also some vessels with no data for several consecutive months.
% Therefore, various operations are performed on the original    data in this article, including cleaning anomalous data, data interpolation, data sampling, and data standardization.
\par %数据处理
% We first purge and segment abnormal data, such as abnormal speed and heading lost data, and data that haven't left the port for a year. For data with time intervals of more than one hour between consecutive track points, we perform segmentation. Then, we adjust the time interval of the trajectory sequence to one minute by independently interpolating the longitude and latitude using cubic spline interpolation. Finally, we use the interpolated latitude and longitude to calculate the sog and cog of various trajectory points, and fill the vessel's heading with the first value.
We first purged the abnormal, such as abnormal speed and heading lost data, and data that haven't left the port for a year. 
For data with time intervals of more than one hour between consecutive track points, we performed segmentation.
Then, we adjusted the time interval of the trajectory sequence to one minute by independently interpolating the longitude and latitude using cubic spline interpolation.
 Next, we used the interpolated latitude and longitude to calculate the sog and cog of various trajectory points and filled the vessel's heading with the first value.
 After the preprocessing, there remain 2378 vessels with 35 types.
 We randomly selected 68 vessels from each type, or all of them if there are not enough.
% First, abnormal data processing mainly includes purging and segmentation. 
% Abnormal speed data and heading lost data are cleared, as well as data that have not left the port for one year. 
% Also, segmentation is performed for data with time intervals of more than one hour between consecutive track points.
% Then, the trajectory sequence's time interval was adjusted to one minute by independently interpolating the longitude and latitude using cubic spline interpolation.
% Finally, Calculate the sog and cog of various trajectory points using the interpolated latitude and longitude, and fill the vessel's heading with the first value.
% After this, a total of 2378 vessels of 35 different types were cleared, with an average of 68 vessels of each type, but there is still an imbalance in \cref{tab:vessel-type}.
% Therefore, this paper randomly selects 68 vessels from each type, or all of them if there are not enough.
Finally, we computed the z-score of the $\Delta sog$ and $\Delta cog$.
% standardization of the $\Delta sog$ and $\Delta cog$ after difference is performed by
% \begin{equation}
% \mathbf{x}_i^{t*}=\frac{\mathbf{x}_i^t-\mu}{\sigma}
% \end{equation}
% where $\mu,\sigma$ is the means and standard deviations of the overall trajectory data.
\par %数据集分割
Four datasets with different prediction durations were generated in \cref{tab:dataset}.
Each dataset consists of three parts of data, including encoded data, decoded data, and prediction data, denoted by $\mathbf{X}_{enc}$, $\mathbf{X}_{dec}$, $\mathbf{Y}$ in Fig. \ref{fig:fig1}.
% Two of the datasets were encoded with a data length of 72, decoded with a length of 48, prediction horizon of 24, and time intervals of 1 minute and 3 minutes, respectively.
% That is, 1.2 hours is used to predict 0.4 hours and 3.6 hours to predict 1.2 hours.
% The length of each part of the other two data sets is 48, 24, and 48, respectively. 
% i.e., 0.8 hours is used to predict 0.8 hours and 2.4 hours is used to predict 2.4 hours.
In addition, We subsetted dataset2 by selecting data points with above-average rates of change of speed and course to create cornerset.
This dataset was specifically used to evaluate the predictive ability of different models for cornering trajectory.
We split each dataset, using 70\% of the data for training, 10\% for validation, and 20\% for testing.
\begin{table}\centering
\caption {Datasets} \label{tab:dataset} 
\resizebox{!}{!}{
\begin{tabular}{cccccc}
\hline
Dataset & \begin{tabular}[c]{@{}c@{}}Encoded\\ Length\end{tabular} & \begin{tabular}[c]{@{}c@{}}Decoded\\ Length\end{tabular} & \begin{tabular}[c]{@{}c@{}}Prediction\\ Horizon\end{tabular} & \begin{tabular}[c]{@{}c@{}}Time\\ Interval\end{tabular} & \multicolumn{1}{c}{Size} \\ \hline
1       & 72(3.6h)                                                  & 48                                                        & 24(1.2h)                                                     & 3 minutes                                               & 43000                    \\
2       & 72(1.2h)                                                  & 48                                                        & 24(0.4h)                                                     & 1 minute                                                & 172100                   \\
3       & 48(2.4h)                                                  & 24                                                        & 48(2.4h)                                                     & 3 minutes                                               & 43000                    \\
4       & 48(0.8h)                                                  & 24                                                        & 48(0.8h)                                                     & 1 minute                                                & 172100                   \\
cornerset       & 72(1.2h)                                                  & 48                                                        & 24(0.4h)                                                     & 1 minute                                                & 32600                   \\    \hline
\end{tabular}}
\end{table}
\subsection{Experimental Settings}

\begin{figure}[htp]
\centering
\includegraphics[width=3.5in]{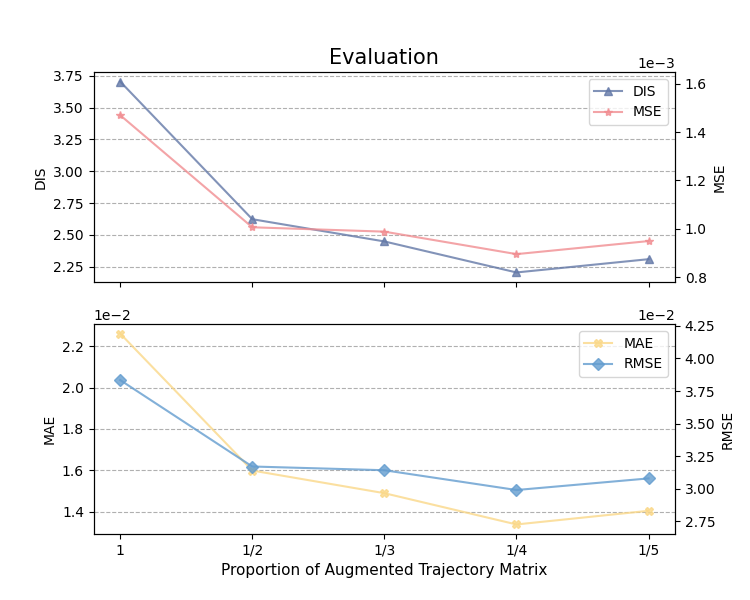}
\caption{Performance of MSTFormer under the different proportions of ATMs}
\label{fig:proportion-of-ATMs}
\end{figure}

\begin{figure}[htp]
\centering
\includegraphics[width=3.5in]{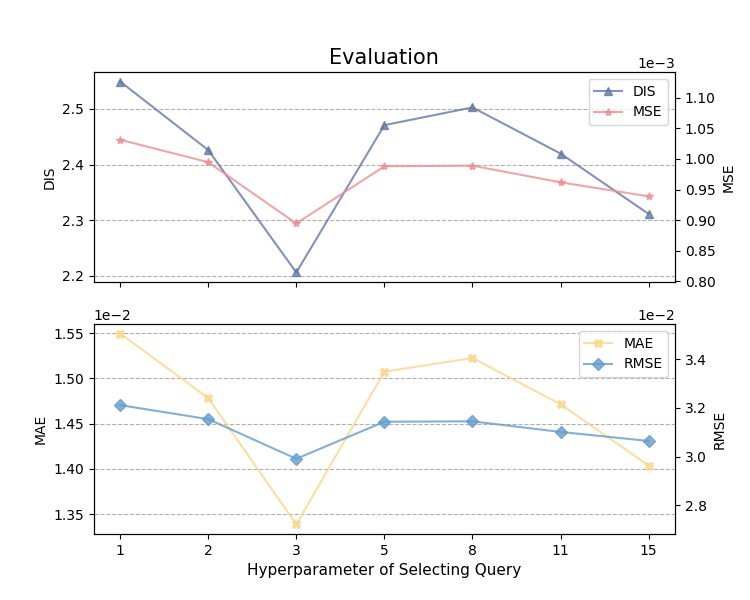}
\caption{Performance of MSTFormer under different number of selected Query}
\label{fig:Hyperparameter-Query}
\end{figure}

Our method aims to improve the long-term vessel trajectory prediction performance of backbone models utilizing motion-inspired model structure and loss function.
So, we compared the proposed MSTFormer with some traditional methods and several state-of-the-art machine learning approaches to illustrate the model structure's superior prediction ability.
Including SVR\cite{drucker1996support}, RFR\cite{breiman2001random}, LSTM, Seq2Seq-LSTM\cite{liu2022deep}, Seq2Seq-LSTM with attention\cite{wang2020long} and Transformer\cite{vaswani2017attention}. 
% Furthermore, We apply our proposed motion-inspired loss function to the deep learning model above to demonstrate its effectiveness.
%方法目的、和哪些方法对比，体现结构优越性，lossfunction的加入，体现loss function对其他方法效果的提升
\par
To ensure the reliability of the results, we used the same training parameters for all deep-learning models.
We set the training epoch to be the ratio of the training dataset length to batch size, meanwhile, the batch size and dropout were set to 50 and 0.05, respectively.
Also, to improve convergence efficiency, the Adam optimizer with 5e-6 learning rate was utilized, which was updated by multiplying $0.5^{epoch-1}$.
In addition, we set up a supervisory mechanism where the model stopped training when the error was larger than the optimal error plus $\alpha$ on the evaluation set, and in this paper $\alpha$ equals 0.5.
All models have three layers, two encoder layers, and one decoder layer.

\par
Two crucial hyperparameters for our model are the quantity of Augmented Trajectory Matrix and the number of selected Query in the dynamic-aware attention since they have the most effects on the prediction results.
In order to select the proper values, we performed experiments:
\textbf{(1)Proportion of Augmented Trajectory Matrix:}
We chose the proportion of ATMs in the series data from $[1,1/2,1/3,1/4,1/5]$ to assess the model's performance.
Fig.\ref{fig:proportion-of-ATMs} shows the DIS, MSE, MAE, and RMSE of the MSTFormer under the different settings.
We can observe that when more data are augmented, it is easy to overfit, leading to worse results.
% It can be seen that when more data are used for augmentation, it leads to an increase in error on the test set due to overfitting. 
% Also when too little data is used, it is difficult to achieve the enhancement effect.
And when the proportion of ATMs is set to $1/4$, MSTFormer performs at its best.
\textbf{(2)Number of Selected Query:}We fixed the proportion of ATMs to 1/4 and selected the hyperparameter $c$ of selected Query from $[1,2,3,5,8,11,15]$, where $c$ was used to control the number of queries that are selected to participate in the computation in \cref{eq:select_Q}.
When the hyperparameter is set to $15$, all 72 variables are involved in the attention calculation.
Fig.\ref{fig:Hyperparameter-Query} shows the DIS, MSE, MAE, and RMSE of the MSTFormer under the different number of selected Query settings.
Our model has the best performance when the hyperparameters are set to 3.

%参数设置、简单参数，以及attention采样的参数图片对比结果。
\subsection{Results and Analysis}
\subsubsection{Prediction Accuracy}

\begin{figure*}[!t]
\centering
\includegraphics[width=\textwidth]{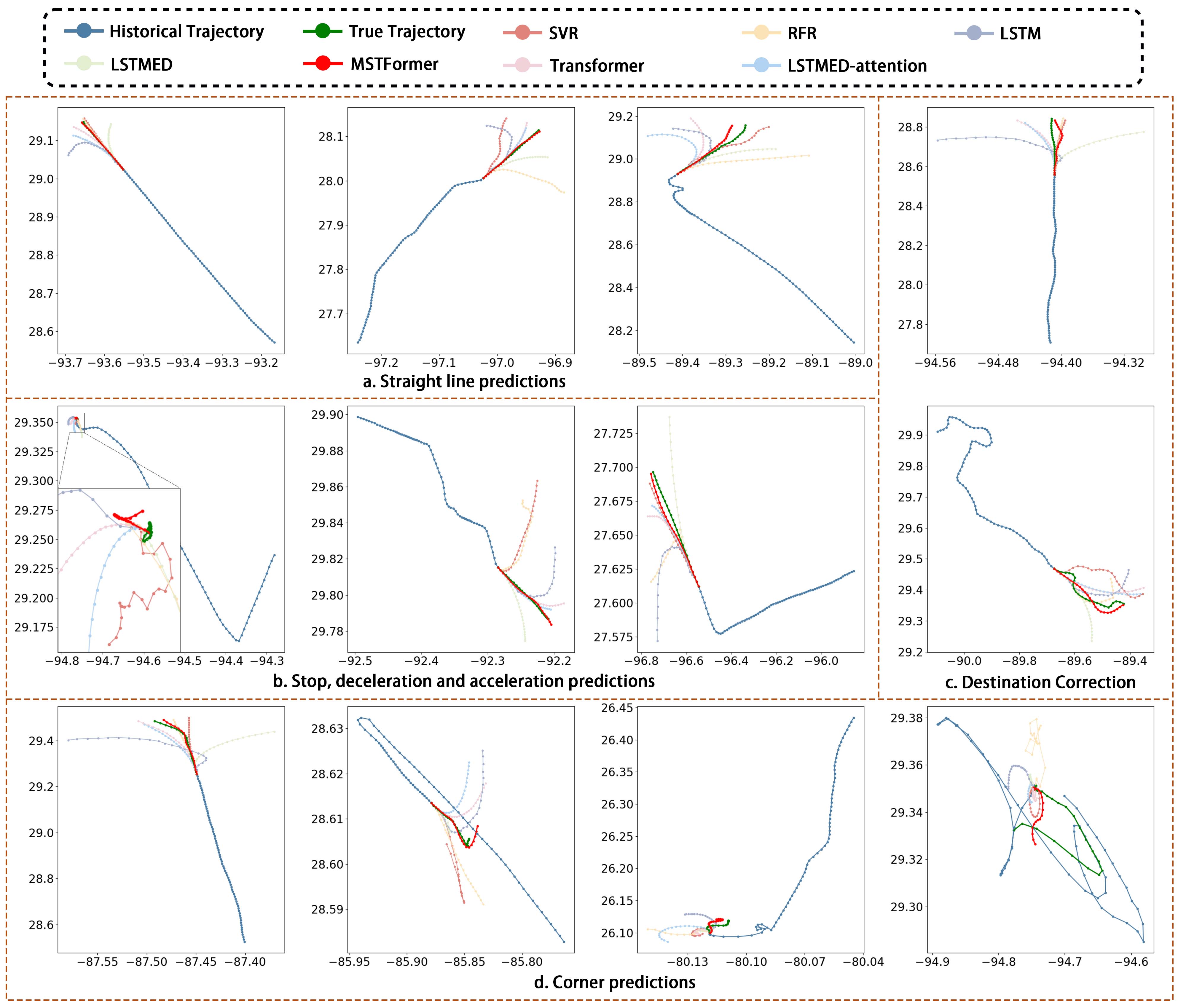}
\caption{Performance of different model}
\label{fig:compare}
\end{figure*}

\begin{table*}\centering
\caption {experimental comparison with baselines and backbones } \label{tab:Accuracy} 
\resizebox{!}{!}{
\begin{tabular}{cccccccc}
\hline
Evaluation matrix                  & SVR    & RFR    & LSTM   & LSTMED & LSTMED-attention & Transformer & MSTFormer \\ \hline
MSE($10^{-3}$) & 1.8779 & 1.6682 & 1.9280 & 1.5328 & 1.2401           & 1.4138      & 0.8946    \\
MAE($10^{-2}$)                     & 1.9213 & 1.9034 & 2.3574 & 2.1339 & 1.6878           & 1.7374      & 1.3384    \\
DIS(KM)                            & 3.1642 & 3.1323 & 3.8756 & 3.4873 & 2.7721           & 2.8634      & \textbf{2.2057}   \\
RMSE($10^{-2}$)                    & 4.3334 & 4.0843 & 4.3909 & 3.9151 & 3.5216           & 3.7600      & 2.9911    \\
MAPE($10^{-4}$)                    & 4.2332 & 4.2131 & 5.3702 & 4.6916 & 3.7678           & 3.8813      & 2.9342    \\
MSPE($10^{-6}$)                    & 1.0996 & 0.9846 & 1.2079 & 0.8296 & 0.7385           & 0.8489      & 0.4793    \\ \hline
\end{tabular}}
\end{table*}
\ 
\newline 
\indent
We compare the performance of the MSTFormer to that of the backbones and baseline approaches using dataset 1.
\cref{tab:Accuracy} shows that the MSTFormer performs better than the traditional methods like SVR and RFR and deep learning methods like LSTM and LSTMED, the former only considers temporal dependencies alone, while the latter is ineffective for long-term cases.
Meanwhile, experiments show that the addition of the self-attention mechanism significantly increases the network's capacity to predict long-term trajectory sequences.
The MSTFormer outperforms its backbones, reducing the DIS by 22.9\% over the Transformer in \cref{tab:Accuracy}, verifying that dynamic knowledge boosts the prediction.

\par
The visual comparisons of various models are shown in Fig.\ref{fig:compare} in order to further assess the prediction capability.
Fig.\ref{fig:compare}(a) demonstrate that the prediction accuracy of different models decreases as the curvature of the historical trajectory increases.
However, the MSTFormer has less influence and performs better.
Meanwhile, MSTFormer has better identification and prediction of vessel's deceleration and stay compared to other models in Fig.\ref{fig:compare}(b).
Fig.\ref{fig:compare}(c) indicate the correction capability of MSTFormer.
It is found that it may be far from the true trajectory in the middle of the prediction horizon but gradually approaches the true value over time.
In addition, MSTFormer also achieves excellent performance on mid-corner and 90-degree turn prediction trajectories In Fig.\ref{fig:compare}(d).
In the last image, although the model accurately predicts the turnaround of the vessel, the subsequent turn prediction is biased. 
However, the final position is very close to the true position, which demonstrates the correction ability of the model.
In summary, when the predicted horizon is more than one hour, the vessels have more freedom and are more difficult to predict. 
And only accurate prediction of speed and course can achieve better performance.

\subsubsection{long-term Prediction}

\begin{table*}\centering
\caption {experimental comparison with different prediction horizons } \label{tab:horizons} 
\resizebox{!}{!}{
\begin{tabular}{ccclcccccc}
\hline
Dataset            & history(h)           & Horizon(h)           & Methods          & MSE($10^{-4}$)   & MAE($10^{-2}$)  & DIS(KM)         & RMSE($10^{-2}$) & MAPE($10^{-4}$) & MSPE($10^{-7}$) \\ \hline
\multirow{3}{*}{1} & \multirow{3}{*}{3.6} & \multirow{3}{*}{1.2} & LSTMED-attention & 12.4018          & 1.6878          & 2.7721          & 3.5216          & 3.7678          & 7.3857          \\
                   &                      &                      & Transformer      & 14.1382          & 1.7374          & 2.8634          & 3.7600          & 3.8813          & 8.4892          \\
                   &                      &                      & MSTFormer        & \textbf{8.9467}  & \textbf{1.3384} & \textbf{2.2057} & \textbf{2.9911} & \textbf{2.9342} & \textbf{4.7934} \\ \hline
\multirow{3}{*}{2} & \multirow{3}{*}{1.2} & \multirow{3}{*}{0.4} & LSTMED-attention & 1.3745           & 0.6382          & 1.0787          & 1.1724          & 1.3983          & 0.7602          \\
                   &                      &                      & Transformer      & 0.9741           & 0.3609          & 0.6453          & 0.9869          & 0.7959          & 0.5356          \\
                   &                      &                      & MSTFormer        & \textbf{0.7411}  & \textbf{0.3212} & \textbf{0.5778} & \textbf{0.8608} & \textbf{0.6956} & \textbf{0.3764} \\ \hline
\multirow{3}{*}{3} & \multirow{3}{*}{2.4} & \multirow{3}{*}{2.4} & LSTMED-attention & 115.9209         & 6.0370          & 9.9655          & 10.7666         & 13.7565         & 74.9649         \\
                   &                      &                      & Transformer      & 247.3624         & 9.4731          & 15.5236         & 15.7277         & 21.9762         & 168.4004        \\
                   &                      &                      & MSTFormer        & \textbf{43.4369} & \textbf{3.2935} & \textbf{5.3971} & \textbf{6.5906} & \textbf{7.2012} & \textbf{24.380} \\ \hline
\multirow{3}{*}{4} & \multirow{3}{*}{0.8} & \multirow{3}{*}{0.8} & LSTMED-attention & 5.5873           & 1.0424          & 1.7310          & 2.3637          & 2.3175          & 3.2291          \\
                   &                      &                      & Transformer      & 5.8840           & 1.1356          & 1.8817          & 2.4257          & 2.5235          & 3.1283          \\
                   &                      &                      & MSTFormer        & \textbf{4.1699}  & \textbf{0.8508} & \textbf{1.4189} & \textbf{2.0420} & \textbf{1.8413} & \textbf{2.1340} \\ \hline
\end{tabular}}
\end{table*}
% \hline
% Dataset & history(h) & Horizon(h) & MSE($10^{-4}$) & MAE($10^{-2}$) & DIS(KM) & RMSE($10^{-2}$) & MAPE($10^{-4}$) & MSPE($10^{-7}$) \\ \hline
% 1       & 3.6            & 1.2        & 8.9467         & 1.3384         & 2.2057 & 2.9911          & 2.9342          & 4.7934          \\
% 2       & 1.2            & 0.4        & 0.7411         & 0.3212         & 0.5778  & 0.8608          & 0.6956          & 0.3764          \\
% 3       & 2.4            & 2.4        & 43.4369         & 3.2935         & 5.3971  & 6.5906          & 7.2012          & 24.380          \\
% 4       & 0.8            & 0.8        & 4.1699         & 0.8508         & 1.4189  & 2.0420          & 1.8413          & 2.1340          \\ \hline

\ 
\newline 
\indent
%随着时间变长预测准确率降低，引用论文xiao2017maritime，说明预测能力，按照图表解释。
We furthermore compare MSTFormer's performance across different prediction horizons; the results are summarized in \cref{tab:horizons}.
The findings indicate that as the prediction horizon lengthens, the predictive accuracy of MSTFormer declines.
It is worth noting that the accuracy of the model improves with the increasing number of observed trajectories.
%不仅随着预测变长，精度变低；观测到的数据越多，预测效果越好，两个方面对比。
In a related study by Xiao et al.\cite{xiao2017maritime}that utilized mined waterway patterns to assist network prediction, the prediction performance was reported to be excellent. 
Specifically, the probability errors for over 50\% of vessels in the 30-minute and 60-minute traffic predictions ranged from 0.6-2 km and 2-4.5 km, respectively. 
However, our study's average error rates for 24-minute and 72-minute traffic predictions were 0.778 km and 2.2057 km for various vessels, respectively.
% the complex method of using the mined waterway patterns to assist network prediction achieved excellent prediction performance.
% In the experimental findings of this study, the probability errors for more than half of the vessels in the 30-minute and 60-minute traffic predictions were between 0.6-2 km and 2-4.5 km, respectively.
% However, the average error of our method for 24-minute and 72-minute traffic prediction is 0.778 km and 2.2057 km for various vessels, respectively.
\par
\begin{figure}[htp]
\centering
\includegraphics[width=3.5in]{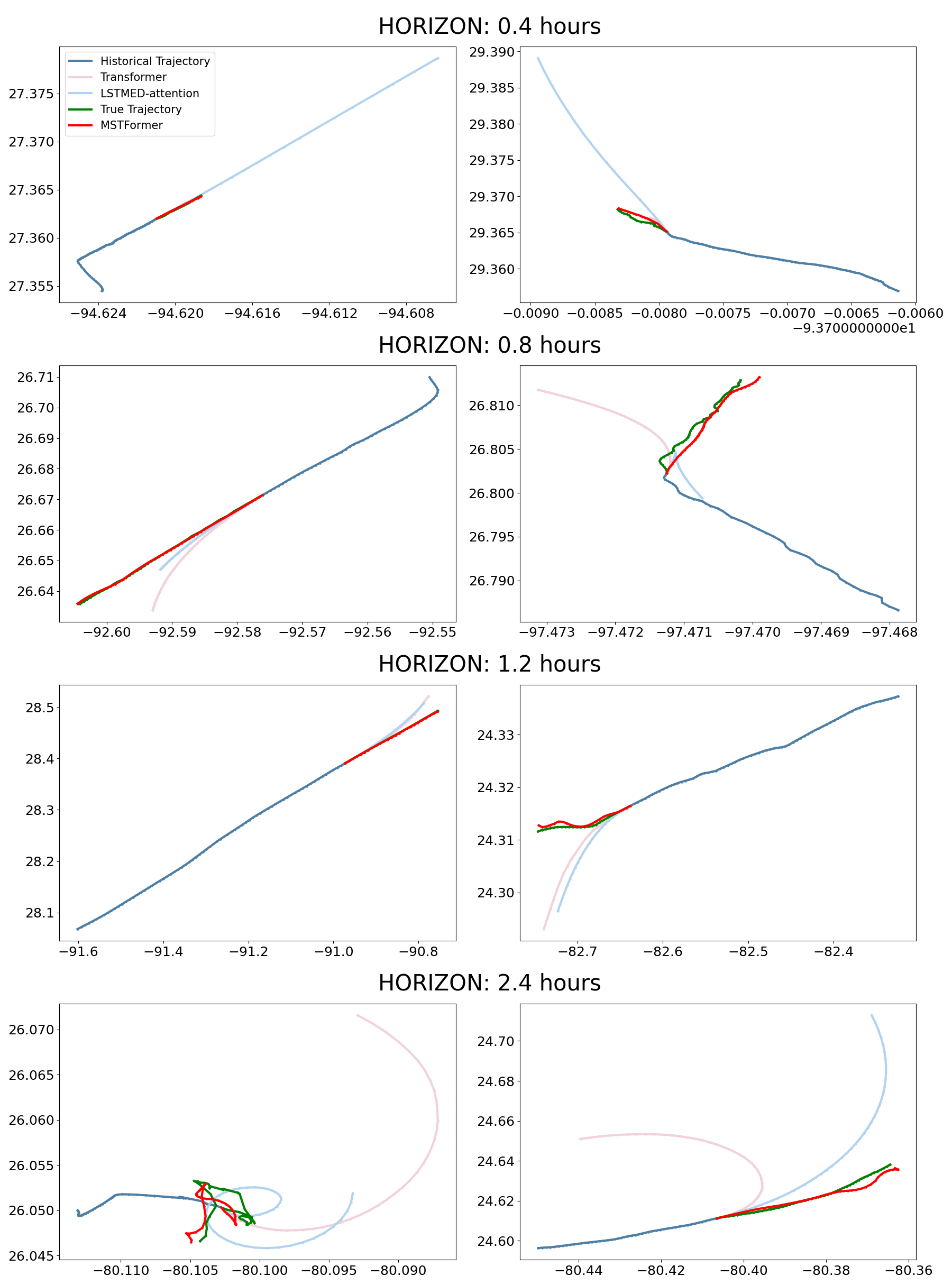}
\caption{Result for different prediction horizon}
\label{fig:allhorizon}
\end{figure}

To explain the results in detail, We have chosen some representative predictions and visualized them in Fig.\ref{fig:allhorizon}.
Our methods demonstrate a high degree of overlap between the true and predicted trajectories when the horizon is 0.4 hours. 
However, as the number of historical trajectories decreases and the prediction horizon increases, the accuracy of our predictions decreases accordingly. 
Despite this, our method still accurately recognizes the vessel's Cornering characteristics in Fig.\ref{fig:allhorizon}. 
As the predicted time continues to increase, the trajectories exhibit increasingly complex navigational behavior, and prediction uncertainty increase. 
Nevertheless, MSTFormer is still able to achieve excellent results in the sixth figure. 
Notably, our model is able to accurately predict the vessels' turns with a final location difference of less than 0.1 KM, as demonstrated in the seventh figure. 
As a result, we conclude that MSTFormer outperforms other models in short and long-term trajectory prediction but struggles in the ultra-long-term trajectory prediction domain (horizons exceeding 1.5h).
% The true and predicted trajectories essentially overlap when the horizon is 0.4 hours.
% When the number of historical trajectory points becomes smaller and the prediction horizon becomes longer, the prediction accuracy decreases subsequently.
% However, the MSTFormer still accurately recognizes the vessel's Cornering characteristics in Fig.\ref{fig:allhorizon}.
% As the predicted time continues to increase, the trajectories contain more complex navigational behavior, and the prediction uncertainty increases, but MSTFormer still achieves good results when the horizon is set to 1.2 hours.
% To our surprise, the model correctly predicts the vessel's turns, and the final location difference is less than 0.1 KM with 1.2 hours prediction horizon in the seventh figure.
% As a result, MSTFormer can retain its dominance in short- and long-term prediction compared to other models, but it faces a significant challenge in ultra-long-term trajectory prediction($>$1.5h).

% \begin{figure}[!t]
% \centering
% \includegraphics[width=3.2in]{picture/experiment2/1.png}
% \caption{Result for prediction horizon of 1.2 hours}
% \label{fig:experiment2-1}
% \end{figure}

% \begin{figure}[!t]
% \centering
% \includegraphics[width=3.2in]{picture/experiment2/2.png}
% \caption{Result for prediction horizon of 0.4 hours}
% \label{fig:experiment2-2}
% \end{figure}

% \begin{figure}[!t]
% \centering
% \includegraphics[width=3.2in]{picture/experiment2/3.png}
% \caption{Result for prediction horizon of 2.4 hours}
% \label{fig:experiment2-3}
% \end{figure}

% \begin{figure}[!t]
% \centering
% \includegraphics[width=3.2in]{picture/experiment2/4.png}
% \caption{Result for prediction horizon of 0.8 hours}
% \label{fig:experiment2-4}
% \end{figure}

\subsubsection{Cornering Trajectory Prediction}
\ 
\newline 
\indent
To test the superiority of the model in predicting the cornering data, we generated cornerset. 
First, we calculated the average speed rate of change and the average course rate of change for dataset 2 as 0.1912 and 4.2746, respectively. 
Then we selected the data in dataset 2 that were larger than the average. 
The final cornerset's average speed change rate and average angle change rate are 0.5539 and 10.9726.
\par
The results show the predicted results of MSTFormer in \cref{tab:cornerset} where DIS is increased from 0.5778 to 1.3781 and MSE is even increased from 0.7411 to 2.5058 compared to \cref{tab:horizons}.
This indicates that compared to the original dataset2, the data in cornerset contains more information and is harder to predict, which is a great challenge for all models.
To observe the predictive ability of other models for cornerset, we selected the backbone for comparison, which performed better in the prediction accuracy experiments in \cref{tab:Accuracy}, namely LSTMED-attention and Transformer.
Surprisingly, the DIS increased from 1.5829 of MSTFormer to 4.9627, a full 213.5\% increase, according to the LSTMED-attention prediction.
Even Transformer increased by 261.5\% compared to MSTFormer.
%ALL_data:172163,sog_average:0.19128841231440893,cog_average:4.274601361773939
%select_data_sog:0.5539077377143505,select_data_cog:10.972667154514168

\begin{table}\centering
\caption{Experimental comparison with different models on cornerset} \label{tab:cornerset}
\resizebox{!}{!}{
\begin{tabular}{cccc}
\hline
\textbf{Metric} & \textbf{LSTMED-attention} & \textbf{Transformer} & \textbf{MSTFormer} \\ \hline
MSE($10^{-4}$) & 24.6290 & 32.7519 & 2.5058 \\
MAE($10^{-2}$) & 2.8034 & 3.2227 & 0.8274 \\
DIS(KM) & 4.5522 & 5.2144 & 1.3781 \\
RMSE($10^{-2}$) & 4.9627 & 5.7229 & 1.5829 \\
MAPE($10^{-4}$) & 5.9454 & 6.8489 & 1.7493 \\
MSPE($10^{-7}$) & 12.7457 & 17.2154 & 1.2272 \\ \hline
\end{tabular}}
\end{table}

% \subsubsection{Knowledge Inspired Loss Function}
\subsubsection{Ablation Study}
\ 
\newline 
\indent
In this subsection, we attempt to evaluate the effectiveness of the individual structures we designed in MSTFormer on the trajectory prediction task.
The prediction results of adding different motion inspired modules we designed to the backbone Transformer are shown in \cref{tab:ablation}.
When we added the information extracted from Augmented Trajectory Matrix using CNN to the network, DIS was reduced from 2.8634 to 2.6331, a reduction of 8.04\%, and MSE was reduced by 12.7\%.
When we change the original attention mechanism to Multi-head Dynamic-aware Self-attention, the prediction error does not decrease but increases a lot.
However, adding the knowledge inspired loss function to the above structure, the prediction error drops sharply, and DIS decreased by 16.2\% from 2.6331 to 2.2057.
To demonstrate the effectiveness of the dynamic-aware attention mechanism, we tried to remove the CNN, and the model still achieved good results. 
This shows that the latter two structures need to be used together to obtain superior performance.
In addition, the results of using the non-corrective loss function improve DIS from 2.2057 to 2.2650.
Overall, the model achieves better prediction accuracy when all structures are used together.

\begin{table*}[!h]\centering
\caption {ablation study of MSTFormer} \label{tab:ablation} 
\resizebox{!}{!}{
\begin{tabular}{cccccccc}
\hline
Evaluation matrix & Transformer & +CNN    & \begin{tabular}[c]{@{}c@{}}+CNN\\ +Dynamic-aware Atten\end{tabular} & MSTFormer       & \begin{tabular}[c]{@{}c@{}}+Dynamic-aware Atten\\ +Knowledge Loss\end{tabular} & \begin{tabular}[c]{@{}c@{}}+CNN\\ +Dynamic-aware Atten\\ +Knowledge Inspird Loss\\ (No correction)\end{tabular} \\ \hline
MSE($10^(-3)$)    & 1.4138      & 1.2336  & 2.4223                                                              & 0.8946          & 0.9287                                                                         & 0.9348                                                                                                          \\
MAE($10^(-2)$)    & 1.7374      & 1.6015  & 2.6690                                                              & 1.3384          & 1.3839                                                                         & 1.3765                                                                                                          \\
DIS(KM)           & 2.8634      & 2.6331  & 4.3762                                                              & \textbf{2.2057} & 2.2741                                                                         & 2.2650                                                                                                          \\
RMSE($10^(-2)$)   & 3.7600      & 3.5122  & 4.9217                                                              & 2.9911          & 3.0475                                                                         & 3.0575                                                                                                          \\
MAPE($10^(-4)$)   & 3.8813      & 3.5384  & 5.9532                                                              & 2.9342          & 3.0212                                                                         & 3.0061                                                                                                          \\
MSPE($10^(-6)$)   & 0.8489      & 0.72032 & 1.3812                                                              & 0.4793          & 0.4918                                                                         & 0.5037                                                                                                          \\ \hline
\end{tabular}}
\end{table*}

\section{Conclusion}\label{sec:conclusions}
In this paper, we propose MSTFormer, the motion inspired transformer, which is incorporated with vessel dynamic knowledge.
% Our method tries to solve the problem of ignoring the correlation between the vessel's own dynamical properties and spatio-temporal characteristics in traditional vessel trajectory prediction methods.
MSTFormer first uses Augmented Trajectory Matrix (ATM) to express the vessel motion state and spatial features. 
Then dynamic-aware self-attention is proposed to enable the backbone Transformer to sense motion changes. 
% by encoding the timestamps to let the network extract temporal features and periodic features. 
Finally, knowledge inspired loss function based on prediction correction is used to train the network to learn Geodesy while understanding the dynamics of the previous construction.
The experimental outcomes prove the superiority of MSTFormer compared with competing methods on normal and corner datasets. 
In addition, the ablation analysis validates the effectiveness of each design module.
\par
% However, our approach is a preliminary effort to apply knowledge of dynamics to capitalize on vessel motion changes and improve prediction performance, there are still some problems and much work to be done in the future.
There are several aspects that are worthy of investigation in future work.
First, ATMs only fuse the motion of vessels; how to incorporate external information, such as coastlines and currents, to achieve better prediction performance is a promising direction.
% First, this paper simply uses a CNN network to extract information from ATMs. 
% Better performance may be obtained with other more sophisticated methods
%When using more sophisticated methods to process our proposed augmented data, it should be possible to extract much richer information.
Second, the ablation analysis shows that only our proposed attention mechanism and the loss function work together to improve the performance of the model, and the reason behind this remains unclear.
%It is also a challenge to make the two methods alone join other networks that can have superior performance.
% Finally, due to the computational complexity of our proposed loss function, how to reduce the training time cost is also a major issue.
Finally, although the proposed model clearly outperforms competing methods on corner data, it remains a challenge to accurately predict trajectories on such data.
\par 
We also anticipate that the idea of incorporating dynamic knowledge with modern neural network architectures sheds light on future research on accurate vessel trajectory prediction.
% \section*{Acknowledgments}

% {\appendix[Proof of the Zonklar Equations]
% Use $\backslash${\tt{appendix}} if you have a single appendix:
% Do not use $\backslash${\tt{section}} anymore after $\backslash${\tt{appendix}}, only $\backslash${\tt{section*}}.
% If you have multiple appendixes use $\backslash${\tt{appendices}} then use $\backslash${\tt{section}} to start each appendix.
% You must declare a $\backslash${\tt{section}} before using any $\backslash${\tt{subsection}} or using $\backslash${\tt{label}} ($\backslash${\tt{appendices}} by itself
%  starts a section numbered zero.)}

\ifCLASSOPTIONcaptionsoff
  \newpage
\fi
\bibliographystyle{ieeetr} %ieeetr国际电气电子工程师协会期刊
\bibliography{ref} % ref就是之前建立的ref.bib文件的前缀

% \newpage

%  the lateX parser from getting confused when it sees the complicated
 

% \begin{IEEEbiography}
% \end{IEEEbiography}



\vfill

\end{document}